\documentclass[sigconf]{acmart}
\usepackage{makecell}
\AtBeginDocument{%
}

\setcopyright{cc}
\setcctype{by}
\copyrightyear{2026}
\acmYear{2026}
\acmDOI{10.1145/3799682.3839938}
\acmConference[CIKM '26]{Proceedings of the 35th ACM International Conference on Information and Knowledge Management}{November 7--11, 2026}{Rome, Italy}
\acmBooktitle{Proceedings of the 35th ACM International Conference on Information and Knowledge Management (CIKM '26), November 7--11, 2026, Rome, Italy}
\acmISBN{979-8-4007-2539-5/2026/11}
\graphicspath{{./}}

\title{Candidate-Fate Accounting for Transparent Sensor Diagnostic Pipeline Search}

\author{Haotao Xie}
\orcid{0009-0002-3061-8090}
\affiliation{%
  \institution{Hangzhou International Innovation Institute, Beihang University}
  \city{Hangzhou}
  \country{China}}
\email{haotaoxie@buaa.edu.cn}

\author{Yutian Chen}
\orcid{0009-0006-6114-8756}
\affiliation{%
  \institution{Hangzhou International Innovation Institute, Beihang University}
  \city{Hangzhou}
  \country{China}}
\email{25601153@buaa.edu.cn}

\author{Yangqi Liu}
\orcid{0009-0003-7688-1190}
\affiliation{%
  \institution{College of Cyber Security, Jinan University}
  \city{Guangzhou}
  \country{China}}
\email{yangqilau@163.com}

\author{Xiaoyu Jiang}
\orcid{0000-0003-4170-5579}
\correspondingauthor
\affiliation{%
  \institution{Hangzhou International Innovation Institute, Beihang University}
  \city{Hangzhou}
  \country{China}}
\email{jiangxiaoyu@buaa.edu.cn}

\renewcommand{\shortauthors}{Xie et al.}

\begin{document}

\begin{abstract}
Industrial sensor diagnostics relies on preprocessing, representation, and classification pipelines, making automated pipeline search useful for reducing manual design cost. However, existing automated machine/deep learning (AutoML/AutoDL) reports typically retain only fitted trials, scores, and winners, omitting generated candidates that are invalid, pruned, skipped, cached, or unfitted. This omission limits reviewers' ability to check signal constraints, budget use, and unevaluated legal alternatives. To address this, we propose candidate-fate accounting, a candidate-level audit framework for diagnostic search traces. It records each observed candidate as auditable evidence: hashes merge repeated observations, legality checks flag invalid candidates, allocation rationales explain budget decisions, and a closed fate ledger assigns one terminal fate to each candidate. Experiments on three bearing-diagnostic datasets show that the framework detects invalid candidates and identifies 30--41 candidates omitted by fitted-trial-only reports, with closed fate records verifying complete candidate accounting while maintaining competitive diagnostic performance. The code is available at \url{https://github.com/XXIE999/candidate-fate-accounting}.
\end{abstract}




\maketitle

\begin{figure}[ht]
\centering
\includegraphics[
    trim=103 55 200 52,
    clip,
    width=0.96\linewidth
]{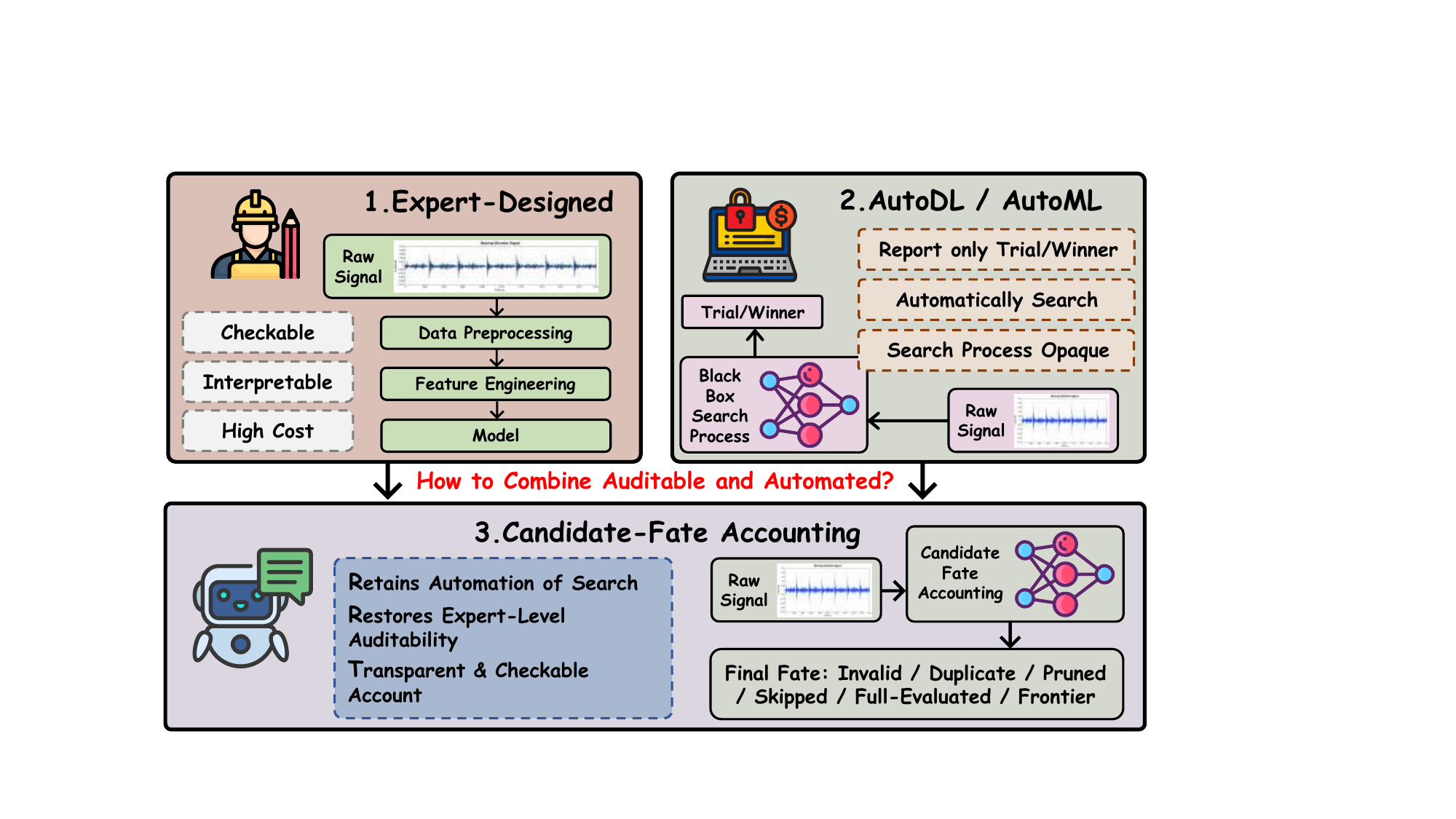}
\Description{A three-part diagram contrasts expert modeling, AutoML/AutoDL search, and candidate-fate accounting. Expert modeling is shown as interpretable and checkable but manually costly; AutoML/AutoDL is shown as automated but limited in search-process visibility; candidate-fate accounting is shown as tracking generated candidates through legality checks, allocation rationales, and final fates.}
\caption{Motivating comparison among expert-designed workflows, AutoML/AutoDL search, and candidate-fate accounting.}
\label{Fig1}
\end{figure}

\section{Introduction}

Industrial sensor diagnostics depends on pipelines that convert machine signals into reliable fault decisions~\cite{Wang2024Rotating}. In deployed maintenance settings, these pipelines must be both accurate and inspectable: engineers need to know which signal transforms are admissible, which classifiers can consume each representation, and why a selected model is plausible~\cite{pr14132112,Shi2024Precision}. Expert-built pipelines provide this reviewability, but the same reasoning is costly to repeat across machines, channels, and operating regimes~\cite{Yang2026Hybrid}. Automated pipeline search~\cite{Mohr2018MLPlanAM,Drori2021AlphaD3MML,MartinSalvador2016AutomaticCA} reduces this manual burden by generating and evaluating candidate preprocessing, representation, and classification pipelines. However, automation should not remove the evidence needed to review the search process. A useful diagnostic search report should show not only which pipeline won, but also what happened to generated candidates that were never fitted. Figure~\ref{Fig1} summarizes this contrast.

Existing work~\cite{2019AutomatedML,Zller2019BenchmarkAS,Mitchell2018ModelCF} addresses parts of this problem but does not close the generated-candidate audit gap. AutoML and AutoDL methods report fitted configurations; hyperparameter-optimization frameworks such as Optuna record scheduled-trial states such as completed, failed, or pruned trials~\cite{akiba2019optuna,Hutter2011SequentialMO,Snoek2012PracticalBO,Falkner2018BOHBRA,Li2018ASF}; grammar and validity methods reject illegal programs; and provenance systems track executed artifacts~\cite{olson2016tpot,feurer2015autosklearn,hadi2023automlbearing,wagner2023sensorless,ferreira2022automlmaintenance,marinescu2021cfgpipeline,nguyen2021avatar,moreau2013prov,Thornton2012AutoWEKACS}. These tools are useful, but their reporting units are usually fitted configurations, scheduled trials, rejected programs, or executed artifacts rather than canonical generated candidates. As a result, repeated proposals, type-invalid candidates that never become trials, and legal candidates skipped by budget or cache decisions may lack one terminal explanation, leaving search validity and budget use difficult to audit at the candidate level. Candidate-fate accounting complements rather than replaces these systems: it adds a canonical candidate-level partition that also covers invalid and non-executed alternatives. This common reporting unit supports reproducible, optimizer-independent comparison of legality and budget traces across AutoML search policies.

A simple diagnostic trace illustrates the gap. Suppose a search generates a short-time Fourier transform (STFT) map with logistic regression, a z-score plus log-mel support vector machine (SVM) pipeline, and a highpass raw-vector XGBoost pipeline, but fits only the third candidate. A fitted-trial report records only the evaluated pipeline and score, hiding that the first is type-invalid and the second is legal but cost-blocked. These hidden outcomes matter under small budgets because they expose legality checks, budget allocation, cache reuse, and untested legal alternatives.

\begin{figure*}[ht]
\centering
\includegraphics[
    trim=5 193 5 10,
    clip,
    width=1.0\textwidth
]{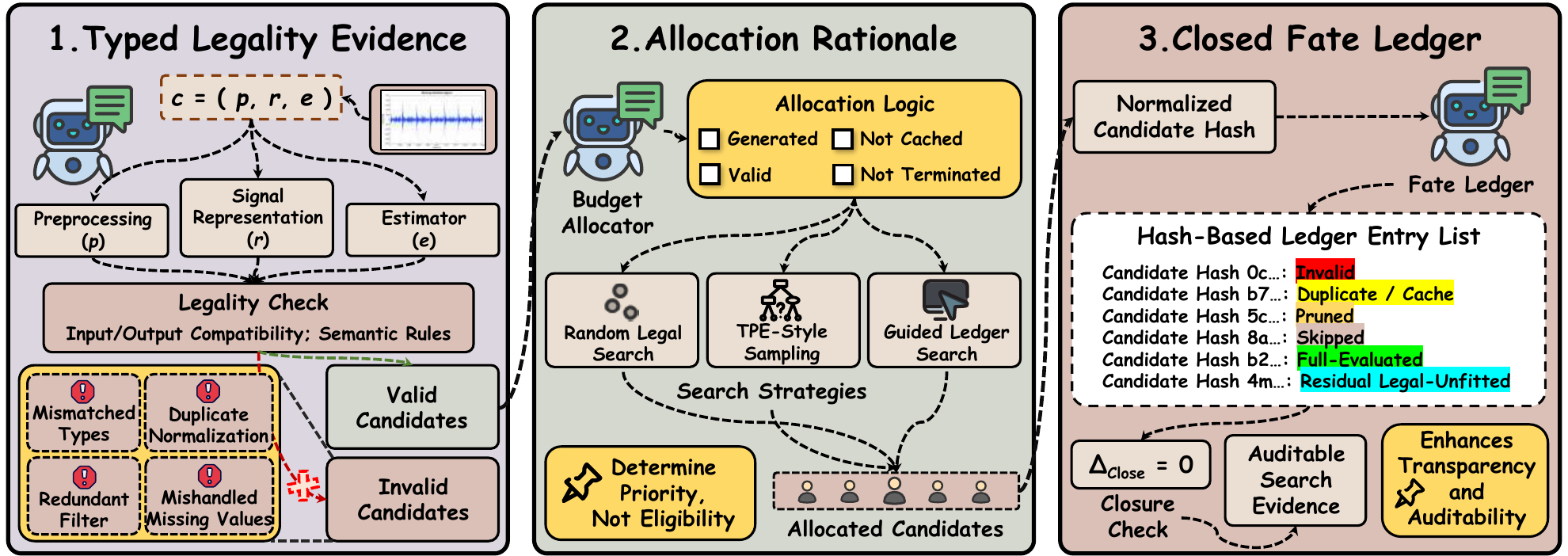}
\Description{A three-stage candidate-fate accounting workflow. First, a generated diagnostic program is decomposed into preprocessing, signal-representation, and estimator components and checked for input/output compatibility and semantic validity. Second, a budget allocator uses replaceable search strategies to prioritize generated, valid, uncached candidates that have not received a terminal fate. Third, normalized candidate hashes merge repeated observations into a ledger that assigns one of six fates: invalid, duplicate or cache, pruned, skipped, full-evaluated, or residual legal-unfitted. A zero closure gap yields auditable search evidence.}
\caption{Overview of candidate-fate accounting. Typed legality checks classify generated candidates; a replaceable allocator prioritizes eligible ones; and a hash-based ledger consolidates repeats and assigns one terminal fate---including non-fitted outcomes---to each observed candidate. The closure condition \(\Delta_{\mathrm{close}}=0\) verifies complete, non-overlapping accounting.}
\label{Fig2}
\end{figure*}

To address this gap, we design candidate-fate accounting, a candidate-level audit framework for emitted diagnostic search traces. The framework closes the record over observed canonical candidates rather than enumerating candidates that never appear. It uses stable hashes to merge repeated observations, typed legality checks to expose type and semantic failures before fitting, allocation rationales to explain budget decisions for legal candidates, and a closed fate ledger with \(\Delta_{\mathrm{close}}\) to verify that each observed candidate receives exactly one terminal fate. Thus, non-fitted candidates become reportable evidence rather than optimizer bookkeeping. Allocation policies remain replaceable: guided ledger search is one allocator designed in our experiments, while candidate-fate accounting is the paper's main contribution. Figure~\ref{Fig2} summarizes the workflow.

We evaluate candidate-fate accounting on three bearing-diagnostic datasets: the Case Western Reserve University (CWRU) bearing dataset, the University of Ottawa bearing dataset (Ottawa), and the Southeast University (SEU) bearing dataset. The experiments test whether the framework detects invalid candidates before fitting, accounts for candidates omitted by fitted-trial-only reports with complete fate accounting, and preserves useful diagnostic performance under a controlled protocol.

The main contributions are:

\textbf{(1)} We define the generated-candidate audit gap in automated industrial diagnostic search, showing why non-fitted candidates should be treated as evidence about validity, budget use, and untested legal alternatives.

\textbf{(2)} We design a typed candidate-fate accounting framework that maps observed canonical candidates to auditable legality, allocation rationale, terminal fate, and closure evidence.

\textbf{(3)} We evaluate the framework on CWRU, Ottawa, and SEU through invalid-candidate probes, closed fate ledgers, allocation checks, and controlled-protocol diagnostic performance.

\section{Method}
\label{sec:method}

We formalize candidate-fate accounting as a reporting contract over emitted diagnostic search traces. As shown in Figure~\ref{Fig2}, the framework records typed legality evidence, allocation rationale, and terminal fates in closed fate ledger. These records make signal constraints, budget use, and unevaluated legal alternatives inspectable while keeping the audit layer optimizer-independent. Section~\ref{sec:typed_space} defines legal diagnostic candidates, Section~\ref{sec:legal_allocator} defines budget allocation, and Section~\ref{sec:closed_contract} assigns final fates.

\subsection{Typed Legality Evidence}
\label{sec:typed_space}

A generated candidate is a diagnostic program \(c=(p,r,e)\), where \(p\) is a preprocessing chain, \(r\) is a signal representation, and \(e\) is an estimator. The objects exchanged by these components have declared types, such as waveform, vector, time--frequency map, and patch sequence. Each primitive \(u\) declares an input type \(\tau_{\mathrm{in}}(u)\), an output type \(\tau_{\mathrm{out}}(u)\), and lightweight metadata for semantic checks. The first legality condition is type compatibility:
\[
\mathcal{C}_{\mathrm{type}}
=\{c=(p,r,e):\tau_{\mathrm{out}}(p)=\tau_{\mathrm{in}}(r),\,
\tau_{\mathrm{out}}(r)=\tau_{\mathrm{in}}(e)\}.
\]
The set \(\mathcal{C}_{\mathrm{type}}\) contains candidates whose adjacent signatures agree. It captures syntactic feasibility before any search policy is considered. This rule turns signal-type constraints into pre-fit evidence by rejecting errors such as feeding a map representation to a vector-only classifier. A deterministic semantic map \(\rho:\mathcal{C}_{\mathrm{type}}\rightarrow\{0,1\}\) resolves conflicts by conservative rejection: a missing-value primitive is invalid when dataset metadata reports zero missingness, while a second z-score normalization or filtering step is invalid as redundant. Fixed rule precedence records one reason when checks overlap.

The second step distinguishes intrinsic legality from budget admissibility. Given budget \(B\) and its cost guard \(\mathrm{CostOK}(c;B)\), we define
\[
\begin{aligned}
\mathrm{Legal}(c) &\equiv c\in\mathcal{C}_{\mathrm{type}}\wedge\rho(c)=1,\\
\mathrm{Admissible}(c;B) &\equiv \mathrm{Legal}(c)\wedge\mathrm{CostOK}(c;B).
\end{aligned}
\]
Thus \(\mathcal{C}_{\mathrm{leg}}=\{c:\mathrm{Legal}(c)\}\) is the legal candidate space independent of budget. Type/semantic-incompatible candidates are invalid before fitting, while legal candidates blocked by cost, cache, or termination remain legal and later receive a non-evaluated fate.

\subsection{Allocation Rationale}
\label{sec:legal_allocator}

Allocation rationale makes budget use reviewable without changing candidate legality. 
At search step \(t\), fitting budget~\cite{Li2016HyperbandAN} may be spent only on candidates that are generated, legal, not cached, and not already assigned a terminal fate. 
Let \(\mathcal{G}_t\) be generated candidates, \(\mathcal{H}_{t-1}\) be cached canonical hashes, \(h(c)\) be the stable hash of candidate \(c\), and \(\ell(c)\) be its current ledger state. The allocatable set is
\[
\mathcal{A}_t=\{c\in\mathcal{G}_t\cap\mathcal{C}_{\mathrm{leg}}:
h(c)\notin\mathcal{H}_{t-1},\ \ell(c)\ \text{unassigned}\}.
\]
Search policies operate only on \(\mathcal{A}_t\): RLS~\cite{bergstra2012random} samples uniformly, the TPE-style sampler~\cite{Bergstra2011AlgorithmsFH} adapts primitive distributions from observed validation macro-F1, and GLS~\cite{Kocsis2006BanditBM} uses MCTS-style selection over grammar-approved macro actions. They change only legal-candidate priority, not legality, budget eligibility, cache handling, or terminal fate.

Optional training-split, label-free descriptors such as impulsiveness, spectral entropy, support, and missingness define \(r_{\mathrm{guide}}\) by adjusting primitive-family priority, but cannot bypass legality or ledger rules.

\begin{table*}[t]
    \caption{Closed \(B=50\) GLS ledger counts under typed legal search, averaged over three seeds.}
\label{tab:explanation_coverage}
\centering
\normalsize
\setlength{\tabcolsep}{0.8pt}
\begin{tabular*}{\textwidth}{@{\extracolsep{\fill}}lcccccccc@{}}
\toprule
Dataset & \makecell{Unique\\Generated} & Invalid & Pruned & Skipped & \makecell{Duplicate/\\Cache} & \makecell{Full\\Evaluated} & \makecell{Residual\\Legal-\\Unfitted} & \makecell{Closure\\Gap} \\
\midrule
CWRU & \(49.7{\pm}0.5\) & \(0{\pm}0\) & \(6.7{\pm}0.9\) & \(23.0{\pm}2.8\) & \(0{\pm}0\) & \(20.0{\pm}2.9\) & \(0{\pm}0\) & 0 \\
Ottawa & \(49.7{\pm}0.5\) & \(0{\pm}0\) & \(20.3{\pm}2.9\) & \(18.7{\pm}4.7\) & \(2.3{\pm}0.9\) & \(8.3{\pm}2.1\) & \(0{\pm}0\) & 0 \\
SEU & \(49.7{\pm}0.5\) & \(0{\pm}0\) & \(6.3{\pm}2.1\) & \(23.7{\pm}3.1\) & \(4.3{\pm}2.1\) & \(15.3{\pm}2.5\) & \(0{\pm}0\) & 0 \\
\bottomrule
\end{tabular*}
\end{table*}

\subsection{Closed Fate Ledger}
\label{sec:closed_contract}

The closed ledger reports the fate of every observed canonical candidate, including candidates that were generated but never fitted. The reporting unit is a canonical hash because one diagnostic program may appear through proposal, cache, probe, or evaluation records. In our implementation, \(h(c)\) is a SHA-256-derived identifier over a deterministic sorted-key serialization of the ordered preprocessing steps, representation, estimator, and their parameters. Each hash bucket stores the serialized candidate, legality result, allocation reason, and observed event types. Repeated proposal, cache, probe, or evaluation records update the existing bucket rather than increasing \(U\). Because component order and parameters are included, the same configuration maps to one identity independently of its event path, while structurally different pipelines remain distinct. Ledger construction therefore separates identity resolution from fate assignment: records are first consolidated by \(h(c)\), precedence is then applied per bucket, and closure is computed over \(U\) unique entries. This prevents repeated observations from inflating generated-candidate counts while preserving the event evidence needed to justify the final state.

Let \(\bar{\mathcal{G}}_T\) be the observed canonical generated set at the end of a run. The fate function \(\ell:\bar{\mathcal{G}}_T\rightarrow\mathcal{S}\) maps each candidate to one state in \(\mathcal{S}=\{s_{\mathrm{inv}},s_{\mathrm{dup}},s_{\mathrm{prn}},s_{\mathrm{skp}},s_{\mathrm{eval}},s_{\mathrm{res}}\}\): invalid, duplicate/cache, pruned, skipped, full-evaluated, or residual legal-unfitted.

Fates are assigned by a fixed precedence order so that each observed canonical candidate contributes to one table cell. The order makes late evidence decisive when a candidate is first proposed cheaply and later fitted. Candidates failing type or semantic checks become invalid. Any candidate that consumes fitting budget becomes full-evaluated, even if earlier records only proposed or probed it. Legal candidates observed only through cache reuse become duplicate/cache. Legal non-duplicates removed by cost, complexity, or multi-fidelity guards become pruned. Legal candidates selected for execution but blocked before fitting become skipped. Remaining legal non-duplicates that are observed but never fitted become residual legal-unfitted.

The closure check tests whether terminal fates form a mutually exclusive and exhaustive partition of the observed set:
\[
\bar{\mathcal{G}}_T=\dot{\bigcup}_{s\in\mathcal{S}}\bar{\mathcal{G}}_T^s,\qquad
\Delta_{\mathrm{close}}=|\bar{\mathcal{G}}_T|-\sum_{s\in\mathcal{S}}|\bar{\mathcal{G}}_T^s|=0.
\]
Here \(\dot{\bigcup}\) denotes a disjoint union, \(\bar{\mathcal{G}}_T^s\) is the subset assigned fate \(s\), and \(\Delta_{\mathrm{close}}\) is zero only when no observed canonical candidate is missing or double-counted.

For each observed canonical candidate, the evidence tuple is
\[
\mathcal{E}(c)=
(r_{\mathrm{type}}(c),r_{\mathrm{guide}}(c),\ell(c)),
\]
where \(r_{\mathrm{type}}\) is the legality rationale, \(r_{\mathrm{guide}}\) is optional allocation metadata or rationale, and \(\ell(c)\) is the terminal fate. The run-level report is
\[
\mathcal{R}_T=(c^\star,\{\mathcal{E}(c):c\in\bar{\mathcal{G}}_T\},B_T,\Delta_{\mathrm{close}}).
\]
Here \(c^\star\) is the selected pipeline and \(B_T=|\bar{\mathcal{G}}^{s_{\mathrm{eval}}}_T|\le B\). This report states which pipeline won, why candidates were ruled out, where budget was spent, and which legal alternatives remained unseen by fitting. For \(N\) trace records, \(U\) unique candidates, and serialized candidate size \(L\), ledger construction costs \(O(NL)\) hashing plus expected \(O(N)\) hash-table updates; the in-memory unique-candidate ledger needs \(O(U)\) storage, and closure is \(O(U)\), with no additional model fitting. Each incoming record requires one hash and an expected \(O(1)\) lookup/update. In our \(B=50\) GLS runs, the main JSON ledger occupies 32.8--37.7 KB per run; exact disk use depends on schema and serialization. At the accounting layer, a new domain supplies a canonical serializer, type/semantic rules, and a mapping from search events to the six terminal fates; hash consolidation, precedence, and closure remain unchanged.

\section{Experiments}
\label{sec:experiments}

We evaluate four research questions (RQs) aligned with the audit framework: whether typed legality exposes invalid candidates before fitting (RQ1), whether the candidate-fate ledger closes over observed generated candidates (RQ2), whether allocation behavior is comparable under the same legal space (RQ3), and whether audited search still returns useful diagnostic pipelines under a controlled protocol (RQ4).

\subsection{Experimental Setup}

\textbf{Protocol.}
We use a controlled small-budget protocol to compare audit counts and diagnostic utility across search policies. For each dataset and seed, all main search conditions share the same typed legal space, primitive cost model, split, and macro-F1 metric, with at most \(B=50\) full model-fitting attempts. This cap applies to fitted candidates, not generated candidates: generated candidates may instead be skipped, pruned, cached, or recorded as residual legal-unfitted. Results are averaged over seeds 42, 43, and 44.

\textbf{Datasets.}
We evaluate three bearing-diagnostic datasets with dataset-appropriate window-level splits. CWRU~\cite{hendriks2022cwru,rosa2024cwru_multilabel,vieira2025realistic} and Ottawa use approximately 60/20/20 train/validation/test splits. SEU uses a cross-condition split \(30\_2\rightarrow20\_0\), with the source condition for training and the target condition split equally for validation and testing.

\subsection{Typed Legality Probe}

RQ1 tests whether typed legality exposes invalid candidates before fitting. We generate 100 weakly constrained skeletons over the shared primitive inventory and label each skeleton with type and semantic checks.

The probe yields 48 type-invalid, 20 semantic-invalid, and 32 legal skeletons. Type failures capture object mismatches such as time--frequency maps followed by vector-only classifiers; semantic failures capture conservative metadata conflicts such as duplicate normalization or repeated filtering. The 68/100 invalid count is a legality-layer stress test, not an estimate of typed main-run invalidity: because the main searches operate within the typed legal space, zero invalid candidates there is expected. Rather than weakening the audit claim, this zero-invalid main-run outcome makes the invariant checkable: the ledger confirms that invalid candidates do not consume fitting budget, and the weak probe identifies the failures that less constrained generation would need to catch and explain.

\subsection{Closed Ledger Accounting}

RQ2 tests whether candidate-fate accounting accounts for emitted candidates dropped by fitted-trial reporting and verifies complete accounting over observed canonical candidates. Each record stores legality evidence, allocation rationale when available, a terminal fate, and a reason. Table~\ref{tab:explanation_coverage} gives a \(B=50\) GLS accounting example aggregated over three seeds.

The main finding is that fitted-trial-only reporting omits substantial trace evidence: the ledger records about 30, 41, and 34 non-full-evaluated canonical candidates on CWRU, Ottawa, and SEU, respectively. The closure check verifies that these fate counts form a complete and non-overlapping partition, with all GLS ledger rows in Table~\ref{tab:explanation_coverage} satisfying \(\Delta_{\mathrm{close}}=0\).

For example, fitted-trial-only reporting on Ottawa exposes only \(8.3{\pm}2.1\) full-evaluated candidates, whereas the ledger attributes \(41.3\) others on average to pruning, skipping, or duplicate/cache. The latter comprise \(20.3\) pruned, \(18.7\) skipped, and \(2.3\) duplicate/cache candidates, so the shortfall from generated candidates to fitted trials becomes attributable rather than unexplained. The interpretability is process-level: legality decisions, candidate fate, and budget consequence, not post-hoc explanations of a fitted model. These fates make the search reviewable but do not rank optimizers.

\subsection{Early Allocation}

RQ3 uses GLS as an allocator case study and compares allocation behavior with the legal space and budget fixed. All rows share typed legality, cache rules, semantic guards, and fitting budget. Best F1 \(\leq 30\) is the best validation macro-F1 within 30 full evaluations; Target Success counts seeds that reach the final RLS validation score, used as a dataset-specific reference target; Evaluations to Target measures the first full-evaluation index at which a run reaches the final RLS validation score, computed only over successful seeds. Low Target Success therefore indicates unstable early allocation.

\begin{table}[ht]
    \caption{Early allocation under a shared typed legal space. Bold marks the best F1 and, among 3/3-success rows, the fewest Evaluations to Target.}
\label{tab:efficiency}
\centering
\normalsize
\setlength{\tabcolsep}{2pt}
\begin{tabular*}{\columnwidth}{@{\extracolsep{\fill}}lcccc@{}}
\toprule
Dataset & Method & \makecell{Best F1\(\leq\)30 \(\uparrow\)} & \makecell{Target Success} & \makecell{Evaluations\\to Target \(\downarrow\)} \\
\midrule
CWRU & RLS & 0.9942\(\pm\)0.0009 & 3/3 & 26.00\(\pm\)7.87 \\
CWRU & TPE & \textbf{0.9982}\(\pm\)0.0009 & 3/3 & 19.00\(\pm\)4.32 \\
CWRU & GLS & \textbf{0.9982}\(\pm\)0.0002 & 3/3 & \textbf{7.00}\(\pm\)\textbf{4.55} \\
\midrule
Ottawa & RLS & 0.9673\(\pm\)0.0063 & 3/3 & 9.00\(\pm\)5.89 \\
Ottawa & TPE & 0.9639\(\pm\)0.0038 & 1/3 & 3.00\(\pm\)0.00 \\
Ottawa & GLS & \textbf{0.9856}\(\pm\)\textbf{0.0026} & 3/3 & \textbf{3.67}\(\pm\)\textbf{1.70} \\
\midrule
SEU & RLS & 0.6751\(\pm\)0.0458 & 3/3 & 28.00\(\pm\)3.56 \\
SEU & TPE & 0.7090\(\pm\)0.0167 & 1/3 & 27.00\(\pm\)0.00 \\
SEU & GLS & \textbf{0.7646}\(\pm\)\textbf{0.0185} & 3/3 & \textbf{7.33}\(\pm\)\textbf{6.85} \\
\bottomrule
\end{tabular*}
\end{table}

\begin{figure}[ht]
\centering
\includegraphics[width=0.95\columnwidth]{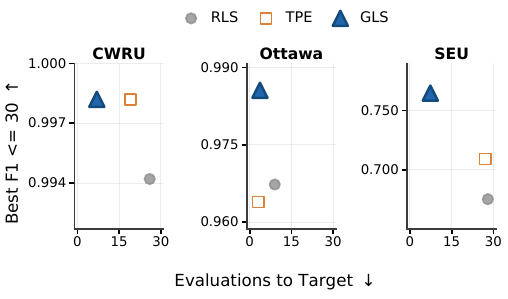}
\caption{Best validation macro-F1 within 30 full evaluations versus Evaluations to Target; upper-left is better.}
\Description{Three panels for CWRU, Ottawa, and SEU plot the best validation macro-F1 within 30 full evaluations against Evaluations to Target. Gray circles denote random legal search, orange squares denote Tree-structured Parzen Estimator-style sampling, and blue triangles denote guided ledger search. Guided ledger search lies in the high-F1, low-evaluation region in all three panels.}
\label{fig:early_allocation_eval_to_target}
\end{figure}

Table~\ref{tab:efficiency} and Figure~\ref{fig:early_allocation_eval_to_target} summarize early-allocation speed and budget-30 utility. GLS gives the most consistent early-allocation profile: it reaches the random-search target in 3/3 seeds on all datasets, requires the fewest Evaluations to Target among 3/3-success rows, attains the best budget-30 F1 on Ottawa and SEU, and ties TPE to four decimals on CWRU. This should be read as an auditability check, not as a broad optimizer ranking: allocation choices can be reported alongside fitted-trial outcomes under the same legal space.

\subsection{Protocol Utility}

RQ4 asks whether the audit framework retains controlled-protocol diagnostic utility. Table~\ref{tab:baseline_aligned} reports controlled-protocol final-test macro-F1 for selected same-space search pipelines, fixed diagnostic recipes, a one-dimensional convolutional neural network (1D-CNN), and AutoML references. Random forest (RF) denotes the estimator used in two fixed recipes. Search rows are matched by split, channel, windowing, label map, metric, and data limit; neural and AutoML rows use different input representations and serve only as scale references.

\balance
\begin{table}[H]
    \caption{Protocol-scoped final-test macro-F1; search rows show three-seed mean (standard deviation), with the best matched search policy per dataset in bold.}
\label{tab:baseline_aligned}
\centering
\normalsize
\setlength{\tabcolsep}{1pt}
\renewcommand{\arraystretch}{0.92}
\begin{tabular*}{\columnwidth}{@{\extracolsep{\fill}}lccc@{}}
\toprule
Method & \makecell{CWRU F1 \(\uparrow\)} & \makecell{Ottawa F1 \(\uparrow\)} & \makecell{SEU F1 \(\uparrow\)} \\
\midrule
RLS & 0.9951(0.0004) & 0.9851(0.0046) & 0.7130(0.0514) \\
TPE & \textbf{0.9978(0.0028)} & 0.9691(0.0119) & 0.7128(0.0242) \\
GLS & 0.9968(0.0000) & \textbf{0.9862(0.0032)} & \textbf{0.7341(0.0134)} \\
\midrule
Statistical + RF & 0.8200 & 0.6216 & 0.0748 \\
Spectral classical & 0.7068 & 0.6181 & 0.0593 \\
Envelope classical & 0.5364 & 0.5926 & 0.5899 \\
Wavelet + RF & 0.8324 & 0.7490 & 0.0901 \\
1D-CNN & 0.7414 & 0.0667 & 0.2380 \\
AutoGluon~\cite{Erickson2020AutoGluonTabularRA} & 0.9172 & 0.8776 & 0.2647 \\
FLAML~\cite{Wang2019FLAMLAF} & 0.9022 & 0.8821 & 0.1371 \\
\bottomrule
\end{tabular*}
\end{table}

Table~\ref{tab:baseline_aligned} shows that accountable search retains controlled-protocol diagnostic utility. Same-space search rows achieve high macro-F1 on CWRU and Ottawa; among matched search policies, TPE is highest on CWRU while GLS is highest on Ottawa and SEU. Because selected pipelines vary across settings, the report records the selected family, transform, estimator, and seed. These results support the scoped claim that candidate-fate accounting can expose terminal fates and allocation outcomes while still selecting plausible final models.

\section{Conclusion}

Candidate-fate accounting addresses the generated-candidate audit gap in automated diagnostic search. By merging repeated trace records, recording legality and allocation rationales, and checking ledger closure with \(\Delta_{\mathrm{close}}\), it turns rejected, skipped, cached, and unfitted candidates into reviewable evidence. Experiments on CWRU, Ottawa, and SEU show that the framework exposes invalid skeletons, attributes non-full-evaluated candidates to explicit fates, and retains useful final-test performance under a controlled protocol. The accounting contract can be adapted through domain-specific candidate schemas and semantic rules, but empirical cross-domain validation remains future work.

\begin{acks}
This work was supported in part by the National Natural Science Foundation of China under Grant 62403425, in part by the Zhejiang Provincial Natural Science Foundation of China under Grant LMS26F030019, in part by the Hangzhou Natural Science Foundation under Grant 2025SZRJJ2330, in part by the Youth Talent Support Project of the Zhejiang Provincial Association for Science and Technology, and in part by the Jiangsu Provincial Scientific Research Center of Applied Mathematics under Grant BK20233002.
\end{acks}

\clearpage

\section*{Generative AI (GenAI) Usage Disclosure}

The authors used generative AI tools in a limited assistive capacity for manuscript language polishing and for code debugging/checking. These tools were not used to generate experimental data, alter results, or make scientific decisions; all code, results, technical claims, and final manuscript text were reviewed and verified by the authors.

\bibliographystyle{ACM-Reference-Format}
\bibliography{references}

@Inbook{olson2016tpot,
author="Olson, Randal S.
and Moore, Jason H.",
editor="Hutter, Frank
and Kotthoff, Lars
and Vanschoren, Joaquin",
title="TPOT: A Tree-Based Pipeline Optimization Tool for Automating Machine Learning",
bookTitle="Automated Machine Learning: Methods, Systems, Challenges",
year="2019",
publisher="Springer International Publishing",
address="Cham",
pages="151--160",
isbn="978-3-030-05318-5",
doi="10.1007/978-3-030-05318-5_8",
url="https://doi.org/10.1007/978-3-030-05318-5_8"
}

@inproceedings{feurer2015autosklearn,
 author = {Feurer, Matthias and Klein, Aaron and Eggensperger, Katharina and Springenberg, Jost and Blum, Manuel and Hutter, Frank},
 booktitle = {Advances in Neural Information Processing Systems},
 editor = {C. Cortes and N. Lawrence and D. Lee and M. Sugiyama and R. Garnett},
 pages = {},
 publisher = {Curran Associates, Inc.},
 title = {Efficient and Robust Automated Machine Learning},
 url = {https://proceedings.neurips.cc/paper_files/paper/2015/file/11d0e6287202fced83f79975ec59a3a6-Paper.pdf},
 volume = {28},
 year = {2015}
}

@article{hadi2023automlbearing,
  title={Improved Fault Classification for Predictive Maintenance in Industrial IoT Based on AutoML: A Case Study of Ball-Bearing Faults},
  author={Russul H. Hadi and Haider Najy Hady and Ahmed Mudheher Hasan and Ammar Abdulhussein Lafta Al-Jodah and Amjad Jaleel Humaidi},
  journal={Processes},
  year={2023},
  url={https://api.semanticscholar.org/CorpusID:258749232}
}

@article{wagner2023sensorless, address={IR},
   title={A framework for the automated parameterization of a sensorless bearing fault detection pipeline},
   volume={10},
   url={https://doi.org/10.22105/jarie.2023.391005.1538},
   DOI={10.22105/jarie.2023.391005.1538},
   number={4},
   journal={Journal of Applied Research on Industrial Engineering},
   publisher={Research Expansion Alliance (REA) on behalf of Ayandegan Institute of Higher Education},
   author={Wagner, Tobias and Gepperth, Alexander and Engels, Elmar},
   year={2023},
   month=Oct }

@article{ferreira2022automlmaintenance,
title = {Using supervised and one-class automated machine learning for predictive maintenance},
journal = {Applied Soft Computing},
volume = {131},
pages = {109820},
year = {2022},
issn = {1568-4946},
doi = {https://doi.org/10.1016/j.asoc.2022.109820},
url = {https://www.sciencedirect.com/science/article/pii/S1568494622008699},
author = {Luís Ferreira and André Pilastri and Filipe Romano and Paulo Cortez}
}

@inproceedings{marinescu2021cfgpipeline,
  title={Searching for Machine Learning Pipelines Using a Context-Free Grammar},
  author={Radu Marinescu and Akihiro Kishimoto and Parikshit Ram and Ambrish Rawat and Martin Wistuba and Paulito Palmes and Adi Botea},
  booktitle={AAAI Conference on Artificial Intelligence},
  year={2021},
  url={https://api.semanticscholar.org/CorpusID:235349043}
}

@article{nguyen2021avatar,
  title={AutoWeka4MCPS-AVATAR: Accelerating Automated Machine Learning Pipeline Composition and Optimisation},
  author={Tien-Dung Nguyen and Bogdan Gabrys and Katarzyna Musial},
  journal={Expert Syst. Appl.},
  year={2020},
  volume={185},
  pages={115643},
  url={https://api.semanticscholar.org/CorpusID:227151926}
}

@article{hendriks2022cwru,
  title={Towards better benchmarking using the CWRU bearing fault dataset},
  author={Jacob Hendriks and Patrick Dumond and David Knox},
  journal={Mechanical Systems and Signal Processing},
  year={2022},
  url={https://api.semanticscholar.org/CorpusID:245603873}
}

@inproceedings{rosa2024cwru_multilabel,
  title={Benchmarking deep learning models for bearing fault diagnosis using the CWRU dataset: A multi-label approach},
  author={Rodrigo Kobashikawa Rosa and Danilo Braga and Danilo Silva},
  year={2024},
  url={https://api.semanticscholar.org/CorpusID:271328200}
}

@article{vieira2025realistic,
  title={Towards a more realistic evaluation of machine learning models for bearing fault diagnosis},
  author={Jo{\~a}o Paulo Vieira and Victor Afonso Bauler and Rodrigo Kobashikawa Rosa and Danilo Silva},
  journal={ArXiv},
  year={2025},
  volume={abs/2509.22267},
  url={https://api.semanticscholar.org/CorpusID:281659355}
}

@inproceedings{moreau2013prov,
  title={PROV-DM: The PROV Data Model},
  author={Khalid Belhajjame and Reza B'Far and James Cheney and Sam Coppens and Stephen Cresswell and Yolanda Gil and Paul Groth and Graham Klyne and Timothy Lebo and Jamie McCusker and Simon Miles and James D. Myers and Satya Sanket Sahoo and Curt Tilmes},
  year={2013},
  url={https://api.semanticscholar.org/CorpusID:65235238}
}

@article{akiba2019optuna,
  title={Optuna: A Next-generation Hyperparameter Optimization Framework},
  author={Takuya Akiba and Shotaro Sano and Toshihiko Yanase and Takeru Ohta and Masanori Koyama},
  journal={Proceedings of the 25th ACM SIGKDD International Conference on Knowledge Discovery \& Data Mining},
  year={2019},
  url={https://api.semanticscholar.org/CorpusID:196194314}
}

@article{Thornton2012AutoWEKACS,
  title={Auto-WEKA: combined selection and hyperparameter optimization of classification algorithms},
  author={Chris J. Thornton and Frank Hutter and Holger H. Hoos and Kevin Leyton-Brown},
  journal={Proceedings of the 19th ACM SIGKDD international conference on Knowledge discovery and data mining},
  year={2012},
  url={https://api.semanticscholar.org/CorpusID:13952689}
}

@article{bergstra2012random,
  title={Random search for hyper-parameter optimization.},
  author={Bergstra, James and Bengio, Yoshua},
  journal={Journal of machine learning research},
  volume={13},
  number={2},
  year={2012}
}

@inproceedings{Bergstra2011AlgorithmsFH,
  title={Algorithms for Hyper-Parameter Optimization},
  author={James Bergstra and R{\'e}mi Bardenet and Yoshua Bengio and Bal{\'a}zs K{\'e}gl},
  booktitle={Neural Information Processing Systems},
  year={2011},
  url={https://api.semanticscholar.org/CorpusID:11688126}
}

@inproceedings{Hutter2011SequentialMO,
  title={Sequential Model-Based Optimization for General Algorithm Configuration},
  author={Frank Hutter and Holger H. Hoos and Kevin Leyton-Brown},
  booktitle={Learning and Intelligent Optimization},
  year={2011},
  url={https://api.semanticscholar.org/CorpusID:6944647}
}

@inproceedings{Snoek2012PracticalBO,
  title={Practical Bayesian Optimization of Machine Learning Algorithms},
  author={Jasper Snoek and H. Larochelle and Ryan P. Adams},
  booktitle={Neural Information Processing Systems},
  year={2012},
  url={https://api.semanticscholar.org/CorpusID:632197}
}

@article{Li2016HyperbandAN,
  title={Hyperband: A Novel Bandit-Based Approach to Hyperparameter Optimization},
  author={Lisha Li and Kevin G. Jamieson and Giulia DeSalvo and Afshin Rostamizadeh and Ameet Talwalkar},
  journal={J. Mach. Learn. Res.},
  year={2016},
  volume={18},
  pages={185:1-185:52},
  url={https://api.semanticscholar.org/CorpusID:11971778}
}

@article{Falkner2018BOHBRA,
  title={BOHB: Robust and Efficient Hyperparameter Optimization at Scale},
  author={Stefan Falkner and Aaron Klein and Frank Hutter},
  journal={ArXiv},
  year={2018},
  volume={abs/1807.01774},
  url={https://api.semanticscholar.org/CorpusID:49571505}
}

@article{Li2018ASF,
  title={A System for Massively Parallel Hyperparameter Tuning},
  author={Liam Li and Kevin G. Jamieson and Afshin Rostamizadeh and Ekaterina Gonina and Jonathan Ben-tzur and Moritz Hardt and Benjamin Recht and Ameet Talwalkar},
  journal={arXiv: Learning},
  year={2018},
  url={https://api.semanticscholar.org/CorpusID:216245794}
}

@article{Mohr2018MLPlanAM,
  title={ML-Plan: Automated machine learning via hierarchical planning},
  author={Felix Mohr and Marcel Wever and Eyke H{\"u}llermeier},
  journal={Machine Learning},
  year={2018},
  volume={107},
  pages={1495-1515},
  url={https://api.semanticscholar.org/CorpusID:51886269}
}

@article{Drori2021AlphaD3MML,
  title={AlphaD3M: Machine Learning Pipeline Synthesis},
  author={Iddo Drori and Yamuna Krishnamurthy and R{\'e}mi Rampin and Raoni Lourenço and Jorge Piazentin Ono and Kyunghyun Cho and Cl{\'a}udio T. Silva and Juliana Freire},
  journal={ArXiv},
  year={2021},
  volume={abs/2111.02508},
  url={https://api.semanticscholar.org/CorpusID:198940685}
}

@article{MartinSalvador2016AutomaticCA,
  title={Automatic Composition and Optimization of Multicomponent Predictive Systems With an Extended Auto-WEKA},
  author={Manuel Martin Salvador and Marcin Budka and Bogdan Gabrys},
  journal={IEEE Transactions on Automation Science and Engineering},
  year={2016},
  volume={16},
  pages={946-959},
  url={https://api.semanticscholar.org/CorpusID:18001834}
}

@article{Erickson2020AutoGluonTabularRA,
  title={AutoGluon-Tabular: Robust and Accurate AutoML for Structured Data},
  author={Nick Erickson and Jonas W. Mueller and Alexander Shirkov and Hang Zhang and Pedro Larroy and Mu Li and Alex Smola},
  journal={ArXiv},
  year={2020},
  volume={abs/2003.06505},
  url={https://api.semanticscholar.org/CorpusID:212725762}
}

@inproceedings{Wang2019FLAMLAF,
  title={FLAML: A Fast and Lightweight AutoML Library},
  author={Chi Wang and Qingyun Wu and Markus Weimer and Erkang Zhu},
  booktitle={Conference on Machine Learning and Systems},
  year={2019},
  url={https://api.semanticscholar.org/CorpusID:229348714}
}

@inproceedings{Kocsis2006BanditBM,
  title={Bandit Based Monte-Carlo Planning},
  author={Levente Kocsis and Csaba Szepesvari},
  booktitle={European Conference on Machine Learning},
  year={2006},
  url={https://api.semanticscholar.org/CorpusID:15184765}
}

@book{2019AutomatedML,
author = {Hutter, Frank and Kotthoff, Lars and Vanschoren, Joaquin},
year = {2019},
month = {01},
pages = {},
title = {Automated Machine Learning - Methods, Systems, Challenges},
isbn = {978-3-030-05317-8},
doi = {10.1007/978-3-030-05318-5}
}

@article{Zller2019BenchmarkAS,
  title={Benchmark and Survey of Automated Machine Learning Frameworks},
  author={Marc-Andr{\'e} Z{\"o}ller and Marco F. Huber},
  journal={J. Artif. Intell. Res.},
  year={2019},
  volume={70},
  pages={409-472},
  url={https://api.semanticscholar.org/CorpusID:210064426}
}

@article{Mitchell2018ModelCF,
  title={Model Cards for Model Reporting},
  author={Margaret Mitchell and Simone Wu and Andrew Zaldivar and Parker Barnes and Lucy Vasserman and Ben Hutchinson and Elena Spitzer and Inioluwa Deborah Raji and Timnit Gebru},
  journal={Proceedings of the Conference on Fairness, Accountability, and Transparency},
  year={2018},
  url={https://api.semanticscholar.org/CorpusID:52946140}
}

@Article{pr14132112,
AUTHOR = {Jiang, Xiaoyu and Xie, Haotao and Wang, Jiayu and Yang, Zeyu and Zhou, Yuanqiang and Yao, Le and Zhu, Zheren},
TITLE = {Agentic AI for Safety-Aware Process Monitoring and Fault Diagnosis: A Review},
JOURNAL = {Processes},
VOLUME = {14},
YEAR = {2026},
NUMBER = {13},
ARTICLE-NUMBER = {2112},
URL = {https://www.mdpi.com/2227-9717/14/13/2112},
ISSN = {2227-9717},
DOI = {10.3390/pr14132112}
}

@article{Wang2024Rotating,
  author  = {Wang, Haitao and Liu, Xiang},
  title   = {Research on Rotating Machinery Fault Diagnosis Based on Improved Multi-target Domain Adversarial Network},
  journal = {Instrumentation},
  year    = {2024},
  volume  = {11},
  number  = {1},
  pages   = {38--50},
  doi     = {10.15878/j.instr.202300151}
}

@article{Shi2024Precision,
  author  = {Shi, Lichen and Guo, Jiahang and Wang, Haitao},
  title   = {A Precision Machining Equipment Fault Diagnosis Based on {CWT} and Improved {ResNeXt}},
  journal = {Instrumentation},
  year    = {2024},
  volume  = {11},
  number  = {2},
  pages   = {36--43},
  doi     = {10.15878/j.instr.202400030}
}

@article{Yang2026Hybrid,
  author  = {Yang, Chen and Yan, Jianwen and Feng, Yixiong and Li, Lei and Tan, Jianrong},
  title   = {Hybrid Deep Learning for Hydraulic Cylinder Fault Diagnosis under Complex Conditions via Multi-Source Signal Fusion},
  journal = {Instrumentation},
  year    = {2026},
  volume  = {13},
  number  = {1},
  pages   = {40--56},
  doi     = {10.15878/j.instr.202600318}
}

\end{document}